\documentclass[10pt,twocolumn,letterpaper]{article}

\usepackage{wacv}     

\definecolor{wacvblue}{rgb}{0.21,0.49,0.74}
\usepackage[pagebackref,breaklinks,colorlinks,allcolors=wacvblue]{hyperref}
\usepackage[T1]{fontenc}
\usepackage{graphicx}
\PassOptionsToPackage{table}{xcolor}
\usepackage{xcolor}
\usepackage{multirow}
\usepackage{amssymb}
\usepackage{tabularx}
\usepackage{pifont}
\usepackage{colortbl}
\usepackage{float}
\usepackage[table]{xcolor}
\usepackage{xfp}
\usepackage{booktabs}
\usepackage{subcaption}
\usepackage{makecell}
\newcommand{\cmark}{\ding{51}} 
\newcommand{\xmark}{\ding{55}} 

\def\wacvPaperID{*****} 
\def\confName{WACV}
\def\confYear{2027}

\title{\LARGE \bf \texttt{Slot-RAE}: Streamlining Object-Centric Learning \\ via Direct Representation Auto-Encoders}

\author{Alexandre Chapin\\
Ecole Centrale de Lyon, LIRIS\\
69130, Ecully, France \\
{\tt\small alexandre.chapin@ec-lyon.fr}
\and
Emmanuel Dellandrea \\
Ecole Centrale de Lyon, LIRIS\\
69130, Ecully, France \\
{\tt\small emmanuel.dellandrea@ec-lyon.fr}
\and
Liming Chen\\
Ecole Centrale de Lyon, LIRIS\\
69130, Ecully, France\\
{\tt\small liming.chen@ec-lyon.fr}
}

\begin{document}
\maketitle
\begin{abstract}
Deploying object-centric models for real-world scene understanding typically requires complex pipelines to achieve both robust scene decomposition and high-fidelity generation. While recent diffusion-based approaches have improved visual quality, they rely heavily on pre-trained generative priors (e.g., Stable Diffusion) and external VAE latent spaces, bypassing the opportunity for direct semantic composition. In this paper, we propose \textbf{Slot-RAE}, a framework that demonstrates true compositionality can be achieved by operating directly within the continuous semantic feature space of visual foundation models (e.g., DINOv3). Slot-RAE leverages a feature-space diffusion process via a Diffusion Transformer (DiT) and a Representation Alignment (REPA) head to compose the VFM latent space directly. Because the generative core is trained from scratch entirely within the frozen VFM space, it eliminates VAE bottlenecks and the need for task-agnostic generative pre-training. Experiments on the COCO dataset confirm that this direct latent composition strategy yields state-of-the-art compositional capabilities. Slot-RAE provides robust zero-shot compositionality, high-fidelity image reconstruction, and effective unsupervised object discovery, functioning much faster and more efficiently than standard object-centric latent diffusion models.
Code and checkpoints are made available at: \url{https://slot-rae.github.io/}

\end{abstract}

\begin{figure*}[t]
    \centering
    \includegraphics[width=0.75\textwidth]{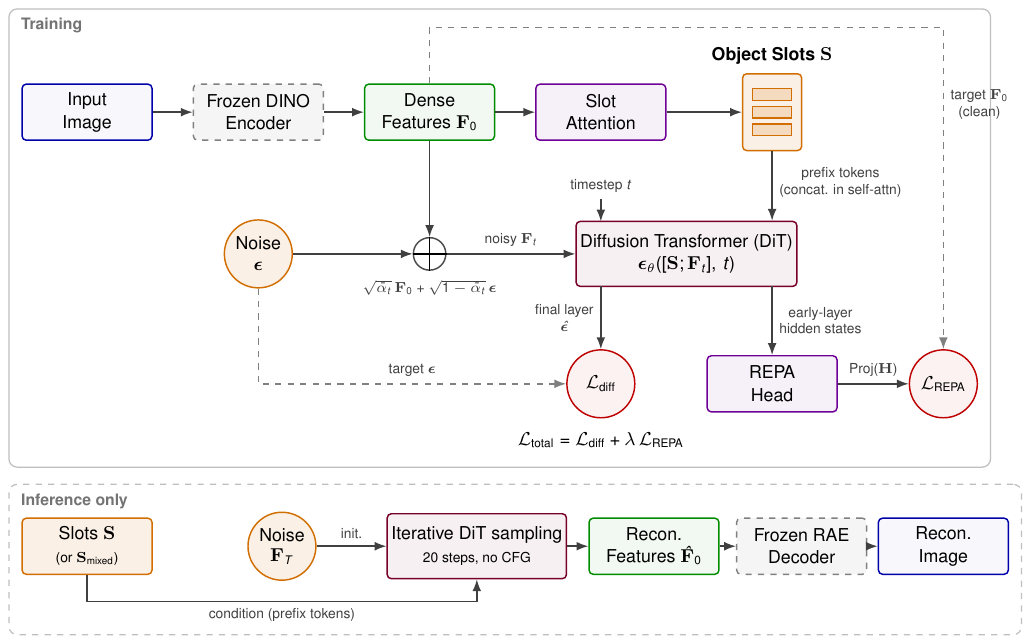}
    \caption{\textbf{Overview of the Slot-RAE Architecture.} The framework processes inputs via a frozen DINO encoder (top row) to extract dense features, which are then compressed into discrete slots via Slot Attention. The generative process (bottom row) diffuses within the continuous DINO feature space utilizing a DiT. The slots condition the diffusion via cross-attention applied to noisy tokens. Both the final DiT layer and the REPA head (branching from an early layer) extract reconstructed features supervised by the target. Finally, an optional, frozen RAE decoder enables zero-shot visualization and latent composition in the pixel space.}
    \label{fig:architecture}
\end{figure*}

\section{Introduction}
\label{sec:intro}

Object-centric learning (OCL) and vision foundation models (VFMs) pursue complementary objectives. The former seeks structured, compositional representations that decompose a visual scene into distinct, interpretable entities~\cite{locatello2020objectcentriclearningslotattention,greff2020bindingproblemartificialneural}, while the latter extracts rich, dense semantic feature spaces from web-scale data~\cite{caron2021emergingpropertiesselfsupervisedvision,oquab2024dinov2learningrobustvisual,Simeoni2025DINOv3}. Recently, the boundaries between these two fields have increasingly blurred: advanced object-centric frameworks now utilize pre-trained VFMs as target representations for reconstruction~\cite{seitzer2023bridginggaprealworldobjectcentric, kakogeorgiou2024spotselftrainingpatchorderpermutation, didolkar2024zeroshotobjectcentricrepresentationlearning}, while foundation models themselves have been shown to exhibit emergent, object-level spatial organization without explicit training for scene decomposition.

To scale OCL to complex, real-world images, the community has progressively transitioned from lightweight spatial broadcast decoders~\cite{watters2019spatialbroadcastdecodersimple} to highly expressive generative architectures~\cite{singh2022illiteratedallelearnscompose,singh2022simpleunsupervisedobjectcentriclearning,zadaianchuk2023objectcentriclearningrealworldvideos, kakogeorgiou2024spotselftrainingpatchorderpermutation}. In particular, recent diffusion-based object-centric methods, such as Slot-Diffusion~\cite{wu2023slotdiffusionobjectcentricgenerativemodeling}, Latent Slot Diffusion~\cite{jiang2023objectcentricslotdiffusion}, GLASS~\cite{singh2025glassguidedlatentslot} and CODA~\cite{nguyen2026improvedobjectcentricdiffusionlearning}, have achieved unprecedented visual fidelity, enabling advanced object-level manipulation and zero-shot compositional generation on complex scenes.


At the same time, Representation Auto-Encoders (RAEs)~\cite{zheng2025diffusiontransformersrepresentationautoencoders, singh2026improvedbaselinesrepresentationautoencoders} have demonstrated that the dense feature spaces of self-supervised foundation models are sufficiently expressive to support high-fidelity generative modeling directly, bypassing traditional VAE compression bottlenecks entirely. While deterministic frameworks like DINOSAUR~\cite{seitzer2023bridginggaprealworldobjectcentric} have successfully established that VFM feature spaces are an excellent substrate for unsupervised object discovery, they lack the capacity for probabilistic synthesis, image manipulation, and compositional generation. Conversely, existing generative diffusion-based OCL methods remain tightly bound to VAE latents and external generative priors. This clear dichotomy prompts a fundamental question: \textit{Can generative object-centric diffusion modeling be formulated directly within the semantic feature space of a foundation model, enabling simultaneous prior-free object discovery and high-fidelity generation?}

In this work, we answer this question affirmatively by introducing \textbf{Slot-RAE}, a novel object-centric generative framework that unifies slot-based scene decomposition with feature-space diffusion. Instead of applying Slot Attention to VAE latents, we perform unsupervised object discovery directly on dense VFM representations and reconstruct these semantic features via a slot-conditioned Diffusion Transformer (DiT)~\cite{peebles2023scalablediffusionmodelstransformers}. To stabilize feature-space generation while preserving fine-grained semantic structures, we integrate an additional Representation Alignment (REPA) objective~\cite{zheng2025diffusiontransformersrepresentationautoencoders}.

Crucially, \textbf{Slot-RAE demonstrates that object-centric diffusion can be trained efficiently on complex, real-world datasets (e.g., COCO~\cite{lin2015microsoftcococommonobjects}) without the large-scale text-to-image priors required by baselines} ~\cite{singh2025glassguidedlatentslot, jiang2023objectcentricslotdiffusion, nguyen2026improvedobjectcentricdiffusionlearning}. The only pre-trained components in our pipeline are the frozen visual backbone defining the semantic feature space and a frozen RAE decoder used strictly for pixel-space visualization and not during training. Operating directly in the semantic feature space yields several key advantages. First, it completely eliminates the structural artifacts and compression bottlenecks imposed by VAEs. Second, it aligns the object discovery mechanism with emergent semantic structures rather than raw pixel statistics, extending the benefits of feature-reconstruction methods to the generative domain. Third, because the reconstructed features can be mapped back to pixel space via a frozen, pre-trained RAE decoder, the synthesis quality remains fully observable while strictly preserving object-centric compositionality.

Extensive experiments on the COCO dataset demonstrate that Slot-RAE effectively bridges the gap between unsupervised object-centric learning and foundation-model-based generative modeling. Our framework achieves competitive unsupervised object discovery performance while producing high-fidelity scene reconstructions and strong zero-shot compositional capabilities. 

Our main contributions are summarized as follows:
\begin{itemize}
    \item We introduce the first object-centric diffusion framework that operates entirely within the semantic feature spaces of vision foundation models, eliminating the need for VAE-based latent representations.
    \item We propose a slot-conditioned Diffusion Transformer with representation alignment that is \textbf{trained entirely from scratch}. This provides the first rigorous diffusion baseline that isolates generative object-centric learning from pre-trained image-generation priors on real-world datasets.
    \item We demonstrate that foundation-model feature spaces naturally support both unsupervised scene decomposition and high-fidelity generation, enabling strong performance on real-world scenes and competitive zero-shot object-level composition.
\end{itemize}


\section{Related Work}
\label{sec:related}

\subsection{Object-centric learning}

Object-centric learning (OCL) aims to resolve the fundamental \emph{binding problem}~\cite{greff2020bindingproblemartificialneural} by decomposing visual scenes into a set of modular entity representations, or slots, without dense supervision. Early approaches, such as MONet~\cite{burgess2019monetunsupervisedscenedecomposition} and Slot Attention~\cite{locatello2020objectcentriclearningslotattention}, demonstrated that object-centric representations could emerge purely through pixel-level reconstruction objectives. However, these methods relied on lightweight decoders, such as Spatial Broadcast Decoders~\cite{watters2019spatialbroadcastdecodersimple}, which severely lack the capacity required to model the appearance complexity of natural images.

To scale OCL to real-world, subsequent works introduced temporal consistency constraints, predictive dynamics, and advanced transformer architectures~\cite{kabra2021simoneviewinvarianttemporallyabstractedobject,elsayed2022saviendtoendobjectcentriclearning,wu2023slotformerunsupervisedvisualdynamics,manasyan2025temporally}. Concurrently, DINOSAUR~\cite{seitzer2023bridginggaprealworldobjectcentric} demonstrated that substituting raw pixel reconstruction with the dense feature spaces of self-supervised VFMs, such as DINO~\cite{caron2021emergingpropertiesselfsupervisedvision} and DINOv2~\cite{oquab2024dinov2learningrobustvisual}, provides a highly robust substrate for object discovery in natural scenes. While existing VFM-based approaches utilize deterministic reconstruction pipelines, our work introduces a probabilistic, generative feature-space formulation via diffusion, unlocking simultaneous unsupervised discovery and high-fidelity compositional synthesis.

\subsection{Diffusion models for object-centric generation}

Diffusion models have established themselves as the dominant paradigm for high-fidelity visual synthesis. Within the OCL domain, they have been adopted to overcome the reconstruction bottlenecks of early spatial decoders and to improve generation quality on complex scenes. Frameworks like Slot-Diffusion~\cite{wu2023slotdiffusionobjectcentricgenerativemodeling}, Latent Slot Diffusion~\cite{jiang2023objectcentricslotdiffusion}, GLASS~\cite{singh2025glassguidedlatentslot}, SlotAdapt~\cite{akan2025slotguidedadaptationpretraineddiffusion} and more recently CODA~\cite{nguyen2026improvedobjectcentricdiffusionlearning} condition diffusion processes on object slots. CODA, in particular, pushes the state-of-the-art in spatial alignment and compositional generation by employing register slots and contrastive alignment within a pre-trained diffusion backbone.

Despite their empirical success, these existing diffusion-based object-centric approaches universally operate within latent spaces obtained from pre-trained VAEs~\cite{kingma2022autoencodingvariationalbayes} and inherit heavily parameterized, pre-trained U-Net diffusion backbones (e.g., Stable Diffusion~\cite{rombach2022highresolutionimagesynthesislatent}). As a result, the learned object representations are tightly coupled with external generative priors. In contrast, Slot-RAE operates directly within the semantic feature space of a VFM and trains its slot-conditioned Diffusion Transformer entirely from scratch, ensuring that the scene decomposition capacity is evaluated without generative pre-training biases.

\subsection{Feature-space generative modeling}

Generative modeling has shifted from pixel-space representations toward semantically abstract latent spaces. Diffusion Transformers (DiTs)~\cite{peebles2023scalablediffusionmodelstransformers} established that transformer-based denoisers offer a highly scalable alternative to traditional convolutional U-Nets. Capitalizing on the semantic richness of vision foundation models, recent approaches have integrated them into generative training in two distinct ways. First, Representation Alignment (REPA)~\cite{yu2025representationalignmentgenerationtraining} demonstrated that aligning the intermediate hidden states of a diffusion model with dense VFM features significantly improves training convergence and semantic consistency. Separately, Representation Autoencoders (RAEs)~\cite{zheng2025diffusiontransformersrepresentationautoencoders, singh2026improvedbaselinesrepresentationautoencoders} proved that the traditional VAE bottleneck can be bypassed completely, showing that diffusion models can be trained to generate directly within the continuous, dense feature spaces of self-supervised foundation models while maintaining high image reconstruction fidelity.

Our work bridges both of these advances with unsupervised object-centric learning. We build upon the RAE framework by replacing the global class or text conditions typical of standard generative models with localized, unsupervised object slots. By training a slot-conditioned DiT to directly denoise foundation-model features from scratch we deliver a unified architecture capable of pure object-centric scene decomposition and generative modeling on complex, real-world distributions.

\section{Methodology}
\label{sec:method}

The Slot-RAE framework, illustrated in Figure~\ref{fig:architecture}, consists of a feature extractor, a slot attention bottleneck, a feature-space diffusion transformer decoder, and an optional frozen RAE image decoder for visual validation.

\subsection{Preliminaries: Slot Attention}
Given an input image $x \in \mathbb{R}^{H \times W \times 3}$, we extract a dense semantic feature map $F_0 \in \mathbb{R}^{h \times w \times d_{in}}$ using a frozen DINOv3~\cite{Simeoni2025DINOv3} (or DINOv2~\cite{oquab2024dinov2learningrobustvisual}) encoder. 

We employ a standard Slot Attention~\cite{locatello2020objectcentriclearningslotattention} module to map these $h \times w$ dense features into a discrete set of $K$ permutation-invariant object tokens, e.g., "slots", $S = \{s_1, s_2, \ldots, s_K\}$, where $s_k \in \mathbb{R}^{d_{slot}}$. The slots are refined through iterative cross-attention, competing to explain different spatial regions of the DINO feature map. For a given iteration $\tau$, the slots are updated via a cross-attention mechanism:
\begin{equation}
S^{(\tau)} = \text{GRU}\left(S^{(\tau-1)}, \text{Softmax}(Q K^T / \sqrt{d_{slot}})V\right)
\end{equation}
where $Q$ is a linear projection of the slots $S^{(\tau-1)}$, and $K, V$ are linear projections of the features $F_0$. Crucially, the normalization inside the softmax is applied over the slots (queries) rather than spatial features (keys). The initial slots $S^{(0)}$ are sampled from a learned Gaussian distribution with parameters $\mu$ and $\Sigma$. The parameters of the Slot Attention module are trained from scratch alongside our decoder.

\subsection{Slot-conditioned feature diffusion}
While prior works reconstruct semantic features, a key contribution of Slot-RAE is its unified generative decoding paradigm. Existing approaches extract slots via VFMs but map them to intermediate VAE latents for diffusion, splitting the latent representation. We resolve this by formulating reconstruction as a conditional diffusion process natively within the dense DINO feature space. This direct modeling eliminates VAE misalignment and unifies extraction and decoding.

\textbf{Diffusion Decoder Transformer (DiT):} We instantiate the decoder as a Diffusion Transformer (DiT)~\cite{peebles2023scalablediffusionmodelstransformers} structured into an encoder-decoder architecture with cross-attention conditioning. We model the generative process using a standard forward diffusion formulation. Given a clean semantic feature map $F_0$, the noisy feature map $F_t$ at diffusion timestep $t$ is defined as:
\begin{equation}
F_t = \sqrt{\bar{\alpha}_t} F_0 + \sqrt{1 - \bar{\alpha}_t} \epsilon, \quad \epsilon \sim \mathcal{N}(0, I)
\end{equation}
where $\bar{\alpha}_t$ is the cumulative product of the noise schedule. 

During training, the DiT receives this noisy target $F_t$ and relies on a hybrid conditioning strategy to guide the denoising process:
\begin{itemize}
\item \textbf{Global modulation (adaLN):} The diffusion timestep $t$ and intermediate noisy spatial features are used to modulate the decoder blocks via standard adaptive layer normalization.
\item \textbf{Object-centric cross-attention:} Unsupervised object slots $S \in \mathbb{R}^{K \times d_{\text{slot}}}$ are extracted directly from the latent features. In the decoder blocks, the spatial tokens of the noisy feature map $F_t$ act as \emph{queries}, while these extracted slots $S$ serve as \emph{keys} and \emph{values} in cross-attention layers.
\end{itemize}
This dual mechanism allows the spatial feature sequence to dynamically pull semantic and structural information from the discovered object slots throughout the denoising process. Finally, we maintain a high-capacity decoder structure~\cite{zheng2025diffusiontransformersrepresentationautoencoders, singh2026improvedbaselinesrepresentationautoencoders} to ensure the network can successfully reconstruct the complex, continuous semantic feature space.

\textbf{Representation Alignment (REPA) Objective:} To stabilize feature-space generation and improve training convergence, we follow~\cite{singh2026improvedbaselinesrepresentationautoencoders} and integrate a Representation Alignment (REPA) objective~\cite{yu2025representationalignmentgenerationtraining}. A REPA projection head is attached to an early-layer output of the DiT to explicitly align the un-noised predicted features with the target DINO space. Let $H_{out}$ represent the targeted DiT hidden state (excluding the prefix slot tokens) at timestep $t$. The REPA loss enforces similarity to the clean semantic features $F_0$:
\begin{equation}
\mathcal{L}_{REPA} = \mathbb{E}_{t, F_0} \left[ \left\| \text{Proj}(H_{out}) - F_0 \right\|_2^2 \right]
\end{equation}

\subsection{Training objective}

While the VFM feature extractor remains strictly frozen, the generative components (the Slot Attention module and the conditional DiT) are trained end-to-end to minimize the noise-prediction objective directly within the continuous, semantic feature space. Let $\epsilon_\theta$ represent our slot-conditioned DiT designed to predict the added noise. The primary diffusion loss is formulated as:
\begin{equation}
\mathcal{L}_{diffusion} = \mathbb{E}_{t, F_0, \epsilon} \left[ \left\| \epsilon - \epsilon_\theta([S; F_t], t) \right\|_2^2 \right]
\end{equation}
To incorporate the representation alignment, the total training objective becomes:
\begin{equation}
\mathcal{L}_{total} = \mathcal{L}_{diffusion} + \lambda \mathcal{L}_{REPA}
\end{equation}
where $\lambda$ is a balancing hyperparameter that is set to 1.0 in our experiments. 

\subsection{Image-space observation and composition}
Because our generated feature map $\hat{F}_0$ is explicitly trained to align with the true VFM distribution, it perfectly matches the input manifold of a pre-trained RAE image decoder~\cite{zheng2025diffusiontransformersrepresentationautoencoders}. We can therefore project $\hat{F}_0$ back to pixel space via $\hat{x} = \text{Decoder}_{RAE}(\hat{F}_0)$, enabling zero-shot observation of the reconstruction quality and visual fidelity.

While traditional approaches like Spatial Broadcast Decoders (SBDs) and autoregressive models can also perform VFM reconstruction, they inherently struggle to support true compositionality. SBDs typically lose fine-grained details and yield excessively noisy reconstructions, whereas autoregressive models tend to overfit to the provided input features. In contrast, operating within our continuous, semantically aligned space unlocks native latent compositionality. By extracting slot representations from multiple, independent images, we can concatenate them into a novel set $S_{mixed}$. Passing this combined slot set through our DiT and the subsequent RAE decoder yields highly coherent, zero-shot compositional scenes. More details are provided in Section~\ref{sec:recons_compos}.

\section{Experiments}
\label{sec:expe}

This section evaluates the hypothesis that object-centric representations learned natively within foundation-model feature spaces can simultaneously achieve accurate scene decomposition and high-fidelity generation without relying on pre-trained generative priors. To validate this hypothesis, we benchmark Slot-RAE on complex, multi-object datasets with diverse visual challenges and evaluation criteria.

\subsection{Experimental Setup}

\paragraph{Datasets.} We evaluate our model on two widely used benchmark datasets for object-centric visual understanding:

\begin{itemize}
    \item \textbf{COCO}~\cite{lin2015microsoftcococommonobjects}: It contains over 200,000 images with more than 80 object categories. The dataset is specifically designed to capture objects in natural, cluttered environments, where multiple instances often appear at different scales and viewpoints. 
    \item \textbf{Pascal VOC}~\cite{everingham2010pascal}: It consists of approximately 11,000 images spanning 20 object categories. Compared with COCO, the dataset contains relatively simpler scenes and fewer object categories, providing a complementary benchmark for evaluating whether learned object representations generalize across diverse semantic concepts.
\end{itemize}

\paragraph{Tasks.} We evaluate Slot-RAE along four complementary dimensions. First, we assess unsupervised \textbf{\textit{object discovery}} by analyzing the spatial masks induced by the slot-attention cross-attention maps. Second, we evaluate  \textbf{\textit{generative synthesis and reconstruction}} by projecting the denoised semantic feature map $\hat{F}_0$ back to pixels via a frozen RAE decoder to measure image fidelity. Third, we test  \textbf{\textit{zero-shot compositional generation}} by removing and swapping slots between independent scenes to demonstrate latent compositionality. Finally, we analyze  \textbf{\textit{computational efficiency}} to quantify the practical benefits of our architecture.

\paragraph{Baselines.} To establish the relative performance of our approach, we compare against representative methods spanning three generations of object-centric learning:
\begin{enumerate*}[label=(\roman*)]
\item pixel reconstruction-based approaches, such as Slot Attention~\cite{locatello2020objectcentriclearningslotattention};
\item deterministic feature-space reconstruction approaches, specifically DINOSAUR~\cite{seitzer2023bridginggaprealworldobjectcentric} with MLP (SBD) and transformer decoders; and
\item diffusion-based generative approaches, including Stable-LSD~\cite{jiang2023objectcentricslotdiffusion}, SlotDiffusion~\cite{wu2023slotdiffusionobjectcentricgenerativemodeling}, GLASS~\cite{singh2025glassguidedlatentslot}, and CODA~\cite{nguyen2026improvedobjectcentricdiffusionlearning}.
\end{enumerate*}
Crucially, these baselines operate under fundamentally different paradigms of supervision. While the first two categories are purely unsupervised, the recent diffusion-based models are inherently weakly-supervised, as they rely heavily on the generative priors of large-scale models (e.g., Stable Diffusion~\cite{rombach2022highresolutionimagesynthesislatent}) trained on massive text-image paired datasets. This broad comparison allows us to isolate the benefits of feature-space diffusion relative to classical unsupervised architectures, while demonstrating that high-fidelity decomposition can be achieved without the burden of large-scale text-to-image priors used by modern generative methods.

In addition to these primary evaluations, we conduct extensive ablation studies, detailed in Appendix~\ref{sec:ablat}, to evaluate the impact of each architectural and design choice.

\subsection{Object discovery}

The primary goal of object discovery is to automatically partition a complex visual scene into its constituent semantic entities without relying on dense mask supervision. To quantify this scene decomposition capability, we evaluate our model using three standard metrics. We report two distinct variants of the Mean Best Overlap (mBO) to capture different levels of grouping: mBO instance ($\text{mBO}_i$), which treats all disconnected components of the same semantic class as separate, independent objects, and mBO class ($\text{mBO}_c$), which groups disconnected components belonging to the same semantic category. We additionally report the Mean Intersection-over-Union ($\text{mIoU}_i$) at the instance level.

\begin{figure}[htbp]
    \centering
    \includegraphics[width=0.11\textwidth]{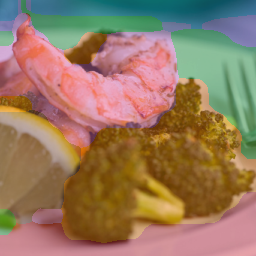} \hfill
    \includegraphics[width=0.11\textwidth]{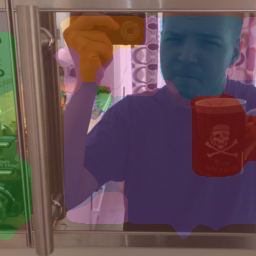} \hfill
    \includegraphics[width=0.11\textwidth]{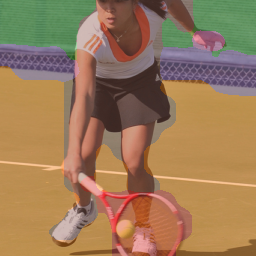} \hfill
    \includegraphics[width=0.11\textwidth]{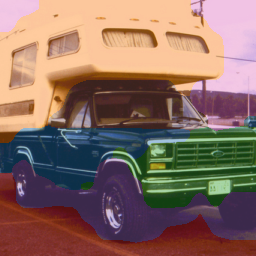}
    
\caption{\textbf{Object discovery} of Slot-RAE on COCO validation images. Each color corresponds to a distinct latent slot, demonstrating the model’s ability to perform unsupervised segmentation of complex scenes.}
    \label{fig:coco_segmentation_results}
\end{figure}

Table~\ref{tab:object_discovery_combined} evaluates whether Slot-RAE preserves the scene decomposition capabilities expected from object-centric learning, comparing it against both unsupervised and weakly-supervised baselines. \textbf{Crucially, our primary objective here is not to establish a new absolute state-of-the-art in pure object discovery. Rather, we aim to demonstrate that operating directly within the continuous VFM latent space for generative modeling does not destroy the model's ability to discover objects, a necessary prerequisite for the latent compositionality demonstrated in subsequent sections.}

First, within the strictly unsupervised regime, we evaluate both DINOv2 and DINOv3 backbone variants for our method and the DINOSAUR baselines to ensure a fair and comprehensive comparison. Interestingly, we observe that the standard MLP and Transformer decoders do not scale well to the more advanced DINOv3 backbone, actually showing a degradation in performance on the COCO dataset. We hypothesize that as the visual foundation model's latent space becomes richer and more structurally complex, simple deterministic decoders struggle to accurately reconstruct the continuous feature manifold from heavily bottlenecked slot representations. In contrast, our DiT-based diffusion decoder models this complex distribution probabilistically, allowing it to fully capitalize on the enhanced semantic features of DINOv3. Across both backbone versions, Slot-RAE consistently outperforms the early pixel-reconstruction approach Slot-Attention, as well as the baseline DINOSAUR (MLP), across all metrics. When compared to the stronger DINOSAUR (Transf.), Slot-RAE achieves competitive object discovery performance on both PASCAL VOC and COCO datasets. Notably, on PASCAL VOC, our DINOv3 variant achieves the highest instance-level segmentation performance (\textbf{49.0} $\text{mIoU}_i$) among all unsupervised baselines. While DINOSAUR acts purely as a non-generative autoencoder, Slot-RAE maintains these strong decomposition capabilities while simultaneously supporting the generative composition of the latent space.

When isolating the comparison to other models capable of generative composition, Slot-RAE demonstrates a clear advantage in scalability. Although SlotDiffusion performs strongly on the PASCAL VOC dataset, Slot-RAE consistently surpasses it across all metrics on the COCO dataset. This indicates that our DiT-based decoder not only preserves high-quality unsupervised discovery capabilities but scales more robustly to complex, real-world scenes than standard pixel-space diffusion decoders.

Finally, we examine the performance gap between Slot-RAE and models that leverage pre-trained text-to-image foundation models, and CODA. While Slot-RAE outperforms the earlier Stable-LSD~\cite{jiang2023objectcentricslotdiffusion} baseline across all metrics (e.g., 32.3 vs. 25.9 $\text{mBO}_i$ on COCO), the more recent GLASS~\cite{singh2025glassguidedlatentslot} and CODA~\cite{nguyen2026improvedobjectcentricdiffusionlearning} models establish a higher performance ceiling. However, this upper-bound gap is fundamentally subsidized by their reliance on the massive, weakly-supervised priors of pre-trained Stable Diffusion. Models like GLASS further inflate their performance with explicit semantic guidance from frozen language captioners. In contrast, Slot-RAE operates in a \textbf{strictly unsupervised regime} under a pure feature reconstruction objective without textual pre-training, mask-guidance scheme, and pre-trained generative U-Nets.

\begin{table*}[htbp]
    \centering
    \small
    \setlength{\tabcolsep}{4pt}
    \caption{\textbf{Object discovery performance.} 
    Comparison across unsupervised and weakly supervised methods. 
    We additionally report whether a method supports compositional generation 
    and its underlying architecture (Backbone / Decoder). Metrics include mBO$_i$ 
    (instance-level), mBO$_c$ (class-level), and mIoU$_i$. 
    \textbf{Bold} indicates the best result and \underline{underline} 
    indicates the second best within each supervision category. Results are reported as a mean of 3 seeds.}
    \label{tab:object_discovery_combined}

    \vspace{0.5em}
    \begin{tabular}{llccccccc} 
        \toprule
        \multirow{2}{*}{\textbf{Model}} 
        & \multirow{2}{*}{\textbf{Comp.}} 
        & \multirow{2}{*}{\textbf{Architecture}} 
        & \multicolumn{3}{c}{\textbf{PASCAL VOC 2012~\cite{everingham2010pascal}}} 
        & \multicolumn{3}{c}{\textbf{COCO~\cite{lin2015microsoftcococommonobjects}}} \\
        \cmidrule(lr){4-6} \cmidrule(lr){7-9} 
        & & & 
        mBO$_i$ & mBO$_c$ & mIoU$_i$ 
        & mBO$_i$ & mBO$_c$ & mIoU$_i$ \\
        \midrule
        
        \multicolumn{9}{l}{\textbf{Unsupervised Object Discovery}} \\ 
        \midrule
        Slot-Attention (MLP)~\cite{locatello2020objectcentriclearningslotattention}
            & \xmark & CNN / MLP
            & 24.6 & 24.9 & --
            & 17.2 & 19.2 & -- \\
        
        DINOSAUR (MLP)~\cite{seitzer2023bridginggaprealworldobjectcentric}
            & \xmark & DINOv2-B / MLP
            & 39.7 & 41.2 & 39.1
            & 28.1 & 32.1 & 26.8 \\
            & \xmark & DINOv3-B / MLP
            & 38.1 & 42.5 & 36.9
            & 25.4 & 31.5 & 24.0 \\
        
        DINOSAUR (Transf.)~\cite{seitzer2023bridginggaprealworldobjectcentric}
            & \xmark & DINOv2-B / Transf.
            & 43.2 & 47.8 & \underline{42.0}
            & \textbf{33.3} & \textbf{41.2} & \textbf{31.6} \\
            & \xmark & DINOv3-B / Transf.
            & 45.2 & 51.9 & 43.8
            & 29.3 & 38.0 & 27.6 \\
        
        SlotDiffusion~\cite{wu2023slotdiffusionobjectcentricgenerativemodeling}
            & \cmark & DINO / U-Net
            & \textbf{50.4} & \textbf{55.3} & --
            & 31.0 & 35.0 & -- \\
        
        \rowcolor{gray!10}
        \textbf{Slot-RAE (Ours)}
            & \cmark & DINOv2-B / \textbf{DiT}
            & 45.0 & 48.6 & 44.1
            & 29.1 & 34.4 & 27.4 \\
        \rowcolor{gray!10}
            & \cmark & DINOv3-B / \textbf{DiT}
            & \underline{50.1} & \underline{52.8} & \textbf{49.0}
            & \underline{32.3} & \underline{38.1} & \underline{30.8} \\
        
        \midrule
        \multicolumn{9}{l}{\textbf{Weakly-Supervised Object Discovery}} \\ 
        \midrule
        Stable-LSD~\cite{jiang2023objectcentricslotdiffusion}
            & \cmark & DINOv2-B / U-Net
            & 32.1 & 35.4 & 31.5
            & 25.9 & 30.0 & 24.7 \\
        CODA~\cite{nguyen2026improvedobjectcentricdiffusionlearning}
            & \cmark & DINOv2-B / U-Net
            & \underline{55.4} & \underline{61.3} & \underline{50.7}
            & \underline{36.6} & \underline{41.4} & \underline{36.4} \\
        GLASS~\cite{singh2025glassguidedlatentslot}
            & \cmark & DINOv2-B / U-Net
            & \textbf{58.9} & \textbf{62.2} & \textbf{58.1}
            & \textbf{40.6} & \textbf{48.5} & \textbf{38.9} \\
        \bottomrule
    \end{tabular}
\end{table*}

\subsection{Image Reconstruction and Zero-Shot Composition}
\label{sec:recons_compos}

Object-centric representations must preserve sufficient scene information while enabling flexible downstream manipulation. To verify these properties, we evaluate our model on two tasks: \textit{image reconstruction} to measure information retention, and \textit{zero-shot composition} to assess the disentanglement and flexibility of the learned slots. We assess structural reconstruction fidelity using PSNR and SSIM~\cite{wangssim}, alongside LPIPS~\cite{zhang2018unreasonableeffectivenessdeepfeatures} for perceptual accuracy. To measure the overall visual realism and distribution alignment of both reconstructed and composed scenes, we use FID~\cite{heusel2018ganstrainedtimescaleupdate} and KID~\cite{binkowski2021demystifyingmmdgans}. Finally, for composition tasks, we use CLIP-I~\cite{ruiz2023dreamboothfinetuningtexttoimage} to quantify how well the semantic identity of an inserted object is preserved from its source image.

\begin{figure}[htbp]
\centering
\includegraphics[width=0.8\linewidth]{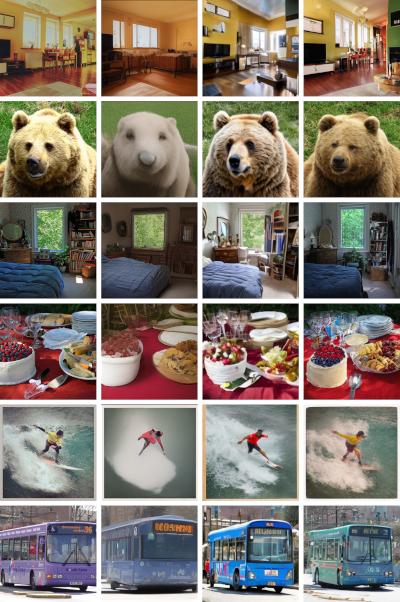}
\caption{\textbf{Qualitative image reconstruction} comparison across
different object-centric models. From left to right: original image, Stable-LSD~\cite{jiang2023objectcentricslotdiffusion}, GLASS~\cite{singh2025glassguidedlatentslot}, and Slot-RAE (ours). The examples span diverse domains including indoor environments, animals, food arrangements, sports scenes, and vehicles.}
\label{fig:model_comparison}
\end{figure}

\paragraph{Image reconstruction}
Figure~\ref{fig:model_comparison} presents qualitative reconstruction results on COCO. Existing methods often struggle to balance fine-grained details with global coherence. Stable-LSD~\cite{jiang2023objectcentricslotdiffusion} tends to oversmooth textures and distort layouts, particularly in dynamic or cluttered scenes. While GLASS~\cite{singh2025glassguidedlatentslot} improves overall composition by leveraging a strong pre-trained Stable Diffusion prior, it frequently hallucinates details or introduces semantic distortions (e.g., altering the cake's content or the surfer's position and color). This suggests its generative prior can bias the solution toward semantically plausible but image-inaccurate reconstructions. In contrast, Slot-RAE produces visually faithful reconstructions with sharp boundaries and preserved textures, closely matching the original scenes without relying on hallucinated priors.

Quantitatively (Table~\ref{tab:combined_metrics}), Slot-RAE consistently outperforms existing diffusion-based baselines across all metrics. Our method achieves the highest structural fidelity (13.93 PSNR, 0.33 SSIM) and lowest perceptual error (0.46 LPIPS, 10.11 FID). This is particularly noteworthy given that \textbf{Slot-RAE is trained entirely from scratch}, without the subsidized generative U-Nets used by GLASS and CODA. Furthermore, Slot-RAE reasonably approaches the theoretical performance upper bound established by the original, non-compositional RAE model~\cite{zheng2025diffusiontransformersrepresentationautoencoders}, indicating minimal information loss during the slot compression process.

\begin{figure*}[t]
    \centering
    
    \begin{subfigure}[b]{0.32\linewidth}
        \centering
        \textbf{Add}\par\medskip
        \includegraphics[width=\linewidth]{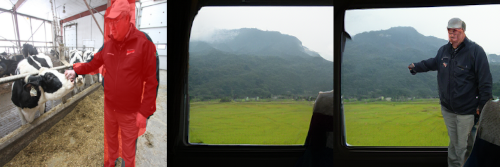}
    \end{subfigure}
    \hfill
    \begin{subfigure}[b]{0.32\linewidth}
        \centering
        \textbf{Subtract}\par\medskip
        \includegraphics[width=\linewidth]{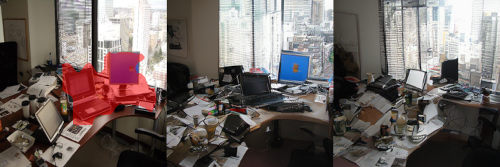}
    \end{subfigure}
    \hfill
    \begin{subfigure}[b]{0.32\linewidth}
        \centering
        \textbf{Compose}\par\medskip
        \includegraphics[width=\linewidth]{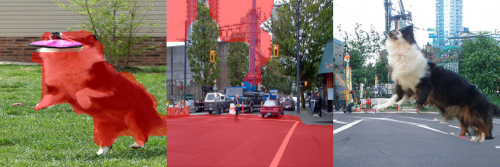}
    \end{subfigure}

    \caption{\textbf{Qualitative evaluation of zero-shot composition.} We evaluate the compositionality of Slot-RAE representations via zero-shot modifications in the semantic feature space. \textit{Left (Addition):} Seamlessly inserts external object slots into a new target context without disrupting the surrounding structural geometry. \textit{Middle (Subtraction):} Removes specific slots from a scene while preserving the global context and filling the occluded space naturally. \textit{Right (Composition):} Blends distinct scene slots across unrelated samples to form a cohesive new scene.}
    \label{fig:slot_algebra}
\end{figure*}

\paragraph{Zero-shot composition}
Following CODA~\cite{nguyen2026improvedobjectcentricdiffusionlearning}, we perform zero-shot composition by swapping, and removing object slots between unrelated COCO images, decoding them without any fine-tuning. Qualitatively (Figure~\ref{fig:slot_algebra}), Slot-RAE preserves object identity and global scene coherence across all operations. As detailed in the ``Composition'' columns of Table~\ref{tab:combined_metrics}, these strengths translate to state-of-the-art quantitative performance. Our method achieves the best scores for global distribution realism (50.04 FID, 50.75 KID) and semantic identity preservation (0.64 CLIP-I). 

\begin{table*}[htbp]
\centering
\small
\setlength{\tabcolsep}{6pt}
\caption{\textbf{Quantitative Evaluation on COCO.} We report image fidelity and global distribution metrics for \textit{reconstruction}, alongside distribution and identity preservation metrics for zero-shot \textit{composition} (slot mixing). Results are mean over 3 random seeds. The best results among object-centric models are in \textbf{bold}, and second best are \underline{underlined}. The Original-RAE is included as a theoretical upper bound for our method on reconstruction but does not support composition natively.}
\label{tab:combined_metrics}
\vspace{0.5em}
\begin{tabular}{l ccccc ccc}
\toprule
\multirow{2}{*}{\textbf{Model}} & \multicolumn{5}{c}{\textbf{Reconstruction}} & \multicolumn{3}{c}{\textbf{Composition}} \\
\cmidrule(lr){2-6} \cmidrule(lr){7-9}
& \textbf{PSNR} $\uparrow$ & \textbf{SSIM} $\uparrow$ & \textbf{LPIPS} $\downarrow$ & \textbf{FID} $\downarrow$ & \textbf{KID ($\times 10^3$)} $\downarrow$ & \textbf{FID} $\downarrow$ & \textbf{KID ($\times 10^3$)} $\downarrow$ & \textbf{CLIP-I} $\uparrow$ \\
\midrule
Stable-LSD~\cite{jiang2023objectcentricslotdiffusion} 
& 10.92 & 0.20 & 0.72 & 125.76 & - & - & - & - \\
GLASS~\cite{singh2025glassguidedlatentslot}
& \underline{11.14} & 0.25 & \underline{0.57} & 21.48 & 6.78 & 55.99 & 51.03 & \underline{0.62} \\
CODA~\cite{nguyen2026improvedobjectcentricdiffusionlearning}
& 10.69 & \underline{0.29} & 0.59 & \underline{13.96} & \underline{2.05} & \underline{53.60} & \underline{53.08} & 0.58 \\
\midrule
\rowcolor{gray!10} \textbf{Slot-RAE (Ours)}
& \textbf{13.93} & \textbf{0.33} & \textbf{0.46} & \textbf{10.11} & \textbf{0.54} & \textbf{50.04} & \textbf{50.75} & \textbf{0.64} \\
\midrule
\textcolor{gray}{\textit{Upper Bound:}} & & & & & & & & \\
Original RAE~\cite{zheng2025diffusiontransformersrepresentationautoencoders} 
& 14.11 & 0.46 & 0.25 & 8.06 & 0.32 & $\emptyset$ & $\emptyset$ & $\emptyset$ \\
\bottomrule
\end{tabular}
\end{table*}

\subsection{Training and inference efficiency}
Compared to existing object-centric diffusion frameworks, our approach bypasses intermediate latent VAEs, large-scale pre-trained diffusion backbones, and legacy U-Net architectures. Instead, our method capitalizes on the recent architectural shift toward Diffusion Transformers (DiT), leveraging their superior scalability and performance~\cite{peebles2023scalablediffusionmodelstransformers}. To quantify the resulting efficiency gains, we evaluate our computational requirements against diffusion-based object-centric baselines. Using an identical hardware configuration, we compare peak GPU memory consumption and single-image inference latency, alongside total parameter counts and training times.

\begin{table}[h]
\centering
\caption{\textbf{Computational Efficiency.} 
Comparison of training requirements, memory usage, and inference latency on a NVIDIA RTX A2000 (8GB) GPU for a single image generation. Lower values indicate higher efficiency.}
\label{tab:efficiency}
\resizebox{\columnwidth}{!}{%
\begin{tabular}{lccccc}
\toprule
& \multicolumn{2}{c}{\textbf{Training Efficiency}} & \multicolumn{2}{c}{\textbf{Inference Efficiency}} \\
\cmidrule(r){2-3} \cmidrule(l){4-5}
\textbf{Model} 
& \textbf{Params (M)} $\downarrow$ 
& \textbf{Train Time} $\downarrow$ 
& \textbf{Memory (GB)} $\downarrow$ 
& \textbf{Latency (s)} $\downarrow$ \\
\midrule
Stable-LSD~\cite{jiang2023objectcentricslotdiffusion} 
    & 1250\textsuperscript{\dag} 
    & 4.0 days 
    & 4.8 
    & 10.3 \\
GLASS~\cite{singh2025glassguidedlatentslot} 
    & 1385\textsuperscript{\dag} 
    & 6.0 days 
    & 5.6 
    & 10.1 \\
CODA~\cite{nguyen2026improvedobjectcentricdiffusionlearning} 
    & 1038\textsuperscript{\dag} 
    & 10.0 days 
    & 4.2
    & 9.4 \\
\midrule
\rowcolor{gray!10}
\textbf{Slot-RAE (Ours)} 
    & \textbf{842}\textsuperscript{\ddag} 
    & \textbf{2.0 days} 
    & \textbf{2.9} 
    & \textbf{1.35} \\
\bottomrule
\multicolumn{5}{l}{\small \textsuperscript{\dag} Includes frozen pre-trained backbone/VAE parameters used during training and also StableDiffusion weights.} \\
\multicolumn{5}{l}{\small \textsuperscript{\ddag} Includes the RAE decoder parameters in addition to the core slot and feature extraction modules.}
\end{tabular}
}
\end{table}

As demonstrated in Table~\ref{tab:efficiency}, our method substantially reduces the computational and structural overhead associated with existing object-centric diffusion pipelines. During training, methods like GLASS necessitate a full image reconstruction pipeline to supervise the generative process. Other baselines, such as CODA and Stable-LSD, operate in a VAE latent space but still require an expensive full forward pass through the heavy U-Net architecture of a pre-trained Stable Diffusion model. In contrast, Slot-RAE shifts the entire diffusion loop natively into a compressed semantic feature space. By utilizing a significantly more compact decoder design and completely bypassing massive pre-trained generative backbones, our approach achieves a two-to fivefold reduction in total training time, converging in just 2.0 days.

At inference, these architectural simplifications yield an even more pronounced efficiency gap, particularly when deployed on resource-constrained hardware such as a consumer laptop GPU (NVIDIA RTX A2000, 8GB VRAM). To ensure a fair comparison, we evaluate all models using an identical budget of 20 denoising steps. Traditional object-centric diffusion frameworks rely heavily on Classifier-Free Guidance (CFG) to strongly bind the generative process to the conditioning slots. While removing CFG in these baselines halves their computational cost, it severely degrades their object-centric generation quality (as demonstrated in CODA's~\cite{nguyen2026improvedobjectcentricdiffusionlearning} Annex), as the models fail to rely adequately on the extracted slots and instead fall back on their unconditioned priors. Consequently, they demand an expensive double forward pass at every iteration to remain effective. In contrast, Slot-RAE performs optimally without CFG, eliminating this overhead entirely. Compounding this advantage, replacing the traditional U-Net with a Diffusion Transformer (DiT) natively aligns with our sequence-based semantic tokens. By stripping away both the CFG requirement and the memory-intensive spatial tensor operations inherent to U-Net skip connections, Slot-RAE slashes single-image inference latency down to a highly responsive 1.35 seconds.


\section{Discussion and limitations}

While Slot-RAE demonstrates that high-fidelity, compositional generation can be achieved without relying on large-scale text-to-image priors, our approach presents several inherent limitations. First, by operating exclusively within the continuous feature space of a VFM, Slot-RAE fundamentally inherits the representational biases and blind spots of its backbone. If the frozen encoder fails to capture specific fine-grained textures, extreme lighting conditions, or exceptionally small objects, these elements remain inaccessible to the diffusion decoder. Consequently, the generative capacity of our model is strictly upper-bounded by the expressivity of the underlying VFM. Furthermore, although our architecture bypasses VAE compression bottlenecks, visualizing the generated scenes in pixel space currently relies on a frozen RAE image decoder. Although this decoder circumvents the need for pixel-level training losses, it may introduce its own visual artifacts during the feature-to-pixel projection, slightly decoupling the measured feature-space accuracy from the final perceptual output. Finally, to bridge the performance gap with weakly-supervised methods, future work could explore integrating text-aligned representation autoencoders, such as the pre-trained models proposed in RAEv2~\cite{singh2026improvedbaselinesrepresentationautoencoders}, to introduce semantic guidance without sacrificing the efficiency of our feature-space formulation.

\section{Conclusion}
In this work, we introduced Slot-RAE, a fully integrated, object-centric generative framework that operates natively within the continuous semantic feature spaces of vision foundation models. By bypassing complex VAE bottlenecks and eliminating the reliance on massive, pre-trained text-to-image priors, we demonstrate that generative object-centric learning can be achieved through a highly streamlined architecture. Utilizing a slot-conditioned Diffusion Transformer alongside a REPA objective, the generative core of Slot-RAE is trained entirely from scratch. This provides a rigorous baseline that successfully isolates scene decomposition capabilities from external generative pre-training biases. Our evaluations on the COCO dataset confirm that this simplified approach yields equivalent or better results than baselines in unsupervised object discovery and outperform even weakly-supervised methods in image-space reconstruction fidelity, and zero-shot compositional generation. Finally, Slot-RAE achieves a subsequent increase in inference speed compared to traditional latent diffusion baselines and substantially reduces training time. Our method establishes a fast, compositionally robust, and highly scalable foundation for bridging the gap between unsupervised scene understanding and controllable visual synthesis.


{
    \small
    \bibliographystyle{ieeenat_fullname}
    \bibliography{main}

\begin{thebibliography}{36}
\providecommand{\natexlab}[1]{#1}
\providecommand{\url}[1]{\texttt{#1}}
\expandafter\ifx\csname urlstyle\endcsname\relax
  \providecommand{\doi}[1]{doi: #1}\else
  \providecommand{\doi}{doi: \begingroup \urlstyle{rm}\Url}\fi

\bibitem[Akan and Yemez(2025)]{akan2025slotguidedadaptationpretraineddiffusion}
Adil~Kaan Akan and Yucel Yemez.
\newblock Slot-guided adaptation of pre-trained diffusion models for
  object-centric learning and compositional generation, 2025.

\bibitem[Bi{\'n}kowski et~al.(2021)Bi{\'n}kowski, Sutherland, Arbel, and
  Gretton]{binkowski2021demystifyingmmdgans}
Miko{\l}aj Bi{\'n}kowski, Danica~J. Sutherland, Michael Arbel, and Arthur
  Gretton.
\newblock Demystifying mmd gans, 2021.

\bibitem[Bock et~al.(2026)Bock, Schüßler, Singh, Schaub-Meyer, and
  Roth]{bock2026mufasamultilayerframeworkslot}
Sebastian Bock, Leonie Schüßler, Krishnakant Singh, Simone Schaub-Meyer, and
  Stefan Roth.
\newblock Mufasa: A multi-layer framework for slot attention, 2026.

\bibitem[Burgess et~al.(2019)Burgess, Matthey, Watters,
  et~al.]{burgess2019monetunsupervisedscenedecomposition}
Christopher~P. Burgess, Loic Matthey, Nicholas Watters, et~al.
\newblock Monet: Unsupervised scene decomposition and representation, 2019.

\bibitem[Caron et~al.(2021)Caron, Touvron, Misra,
  et~al.]{caron2021emergingpropertiesselfsupervisedvision}
Mathilde Caron, Hugo Touvron, Ishan Misra, et~al.
\newblock Emerging properties in self-supervised vision transformers.
\newblock In \emph{ICCV}, 2021.

\bibitem[Didolkar et~al.(2024)Didolkar, Zadaianchuk, Goyal,
  et~al.]{didolkar2024zeroshotobjectcentricrepresentationlearning}
Aniket Didolkar, Andrii Zadaianchuk, Anirudh Goyal, et~al.
\newblock Zero-shot object-centric representation learning, 2024.

\bibitem[Elsayed et~al.(2023)Elsayed, Mahendran, van Steenkiste,
  et~al.]{elsayed2022saviendtoendobjectcentriclearning}
Gamaleldin~F. Elsayed, Aravindh Mahendran, Sjoerd van Steenkiste, et~al.
\newblock Savi++: Towards end-to-end object-centric learning from real-world
  videos.
\newblock In \emph{ICLR}, 2023.

\bibitem[Everingham et~al.(2010)Everingham, Van~Gool, Williams, Winn, and
  Zisserman]{everingham2010pascal}
Mark Everingham, Luc Van~Gool, Christopher K.~I. Williams, John Winn, and
  Andrew Zisserman.
\newblock The pascal visual object classes (voc) challenge.
\newblock \emph{IJCV}, 88\penalty0 (2):\penalty0 303--338, 2010.

\bibitem[Greff et~al.(2020)Greff, van Steenkiste, and
  Schmidhuber]{greff2020bindingproblemartificialneural}
Klaus Greff, Sjoerd van Steenkiste, and Jürgen Schmidhuber.
\newblock On the binding problem in artificial neural networks, 2020.

\bibitem[Heusel et~al.(2017)Heusel, Ramsauer, Unterthiner, Nessler, and
  Hochreiter]{heusel2018ganstrainedtimescaleupdate}
Martin Heusel, Hubert Ramsauer, Thomas Unterthiner, Bernhard Nessler, and Sepp
  Hochreiter.
\newblock Gans trained by a two time-scale update rule converge to a local nash
  equilibrium.
\newblock In \emph{NeurIPS}, 2017.

\bibitem[Jiang et~al.(2023)Jiang, Deng, Singh,
  et~al.]{jiang2023objectcentricslotdiffusion}
Jindong Jiang, Fei Deng, Gautam Singh, et~al.
\newblock Object-centric slot diffusion.
\newblock In \emph{NeurIPS}, 2023.

\bibitem[Kabra et~al.(2021)Kabra, Zoran, Erdogan,
  et~al.]{kabra2021simoneviewinvarianttemporallyabstractedobject}
Rishabh Kabra, Daniel Zoran, Goker Erdogan, et~al.
\newblock Simone: View-invariant, temporally-abstracted object representations
  via unsupervised video decomposition.
\newblock In \emph{NeurIPS}, 2021.

\bibitem[Kakogeorgiou et~al.(2024)Kakogeorgiou, Gidaris, Karantzalos, and
  Komodakis]{kakogeorgiou2024spotselftrainingpatchorderpermutation}
Ioannis Kakogeorgiou, Spyros Gidaris, Konstantinos Karantzalos, and Nikos
  Komodakis.
\newblock Spot: Self-training with patch-order permutation for object-centric
  learning with autoregressive transformers.
\newblock In \emph{CVPR}, 2024.

\bibitem[Kingma and Welling(2014)]{kingma2022autoencodingvariationalbayes}
Diederik~P Kingma and Max Welling.
\newblock Auto-encoding variational bayes.
\newblock In \emph{ICLR}, 2014.

\bibitem[Lin et~al.(2014)Lin, Maire, Belongie,
  et~al.]{lin2015microsoftcococommonobjects}
Tsung-Yi Lin, Michael Maire, Serge Belongie, et~al.
\newblock Microsoft coco: Common objects in context.
\newblock In \emph{ECCV}, 2014.

\bibitem[Locatello et~al.(2020)Locatello, Weissenborn, Unterthiner,
  et~al.]{locatello2020objectcentriclearningslotattention}
Francesco Locatello, Dirk Weissenborn, Thomas Unterthiner, et~al.
\newblock Object-centric learning with slot attention.
\newblock In \emph{NeurIPS}, 2020.

\bibitem[Manasyan et~al.(2025)Manasyan, Seitzer, Radovic,
  et~al.]{manasyan2025temporally}
Anna Manasyan, Maximilian Seitzer, Filip Radovic, et~al.
\newblock Temporally consistent object-centric learning by contrasting slots.
\newblock In \emph{CVPR}, 2025.

\bibitem[Nguyen et~al.(2026)Nguyen, Takida, Murata, Lai, Uesaka, Ermon, and
  Mitsufuji]{nguyen2026improvedobjectcentricdiffusionlearning}
Bac Nguyen, Yuhta Takida, Naoki Murata, Chieh-Hsin Lai, Toshimitsu Uesaka,
  Stefano Ermon, and Yuki Mitsufuji.
\newblock Improved object-centric diffusion learning with registers and
  contrastive alignment, 2026.

\bibitem[Oquab et~al.(2024)Oquab, Darcet, Moutakanni,
  et~al.]{oquab2024dinov2learningrobustvisual}
Maxime Oquab, Timothée Darcet, Théo Moutakanni, et~al.
\newblock Dinov2: Learning robust visual features without supervision, 2024.

\bibitem[Peebles and Xie(2023)]{peebles2023scalablediffusionmodelstransformers}
William Peebles and Saining Xie.
\newblock Scalable diffusion models with transformers.
\newblock In \emph{ICCV}, 2023.

\bibitem[Rombach et~al.(2022)Rombach, Blattmann, Lorenz, Esser, and
  Ommer]{rombach2022highresolutionimagesynthesislatent}
Robin Rombach, Andreas Blattmann, Dominik Lorenz, Patrick Esser, and Björn
  Ommer.
\newblock High-resolution image synthesis with latent diffusion models.
\newblock In \emph{CVPR}, 2022.

\bibitem[Ruiz et~al.(2023)Ruiz, Li, Jampani, Pritch, Rubinstein, and
  Aberman]{ruiz2023dreamboothfinetuningtexttoimage}
Nataniel Ruiz, Yuanzhen Li, Varun Jampani, Yael Pritch, Michael Rubinstein, and
  Kfir Aberman.
\newblock Dreambooth: Fine tuning text-to-image diffusion models for
  subject-driven generation, 2023.

\bibitem[Seitzer et~al.(2023)Seitzer, Horn, Zadaianchuk,
  et~al.]{seitzer2023bridginggaprealworldobjectcentric}
Maximilian Seitzer, Max Horn, Andrii Zadaianchuk, et~al.
\newblock Bridging the gap to real-world object-centric learning.
\newblock In \emph{ICLR}, 2023.

\bibitem[Sim{\'e}oni et~al.(2025)Sim{\'e}oni, Vo, Seitzer, Baldassarre, Oquab,
  Jose, Khalidov, Szafraniec, Yi, Ramamonjisoa, Massa, Haziza, Wehrstedt, Wang,
  Darcet, Moutakanni, Sentana, Roberts, Vedaldi, Tolan, Brandt, Couprie,
  Mairal, J{\'e}gou, Labatut, and Bojanowski]{Simeoni2025DINOv3}
Oriane Sim{\'e}oni, Huy~V. Vo, Maximilian Seitzer, Federico Baldassarre, Maxime
  Oquab, Cijo Jose, Vasil Khalidov, Marc Szafraniec, Seungeun Yi, Michael
  Ramamonjisoa, Francisco Massa, Daniel Haziza, Luca Wehrstedt, Jianyuan Wang,
  Timoth{\'e}e Darcet, Th{\'e}o Moutakanni, Leonel Sentana, Claire Roberts,
  Andrea Vedaldi, Jamie Tolan, John Brandt, Camille Couprie, Julien Mairal,
  Herv{\'e} J{\'e}gou, Patrick Labatut, and Piotr Bojanowski.
\newblock Dinov3, 2025.

\bibitem[Singh et~al.(2022{\natexlab{a}})Singh, Deng, and
  Ahn]{singh2022illiteratedallelearnscompose}
Gautam Singh, Fei Deng, and Sungjin Ahn.
\newblock Illiterate dall-e learns to compose.
\newblock In \emph{ICLR}, 2022{\natexlab{a}}.

\bibitem[Singh et~al.(2022{\natexlab{b}})Singh, Wu, and
  Ahn]{singh2022simpleunsupervisedobjectcentriclearning}
Gautam Singh, Yi-Fu Wu, and Sungjin Ahn.
\newblock Simple unsupervised object-centric learning for complex and
  naturalistic videos.
\newblock In \emph{NeurIPS}, 2022{\natexlab{b}}.

\bibitem[Singh et~al.(2026)Singh, Zheng, Wu, Zhang, Shechtman, and
  Xie]{singh2026improvedbaselinesrepresentationautoencoders}
Jaskirat Singh, Boyang Zheng, Zongze Wu, Richard Zhang, Eli Shechtman, and
  Saining Xie.
\newblock Improved baselines with representation autoencoders, 2026.

\bibitem[Singh et~al.(2025)Singh, Schaub-Meyer, and
  Roth]{singh2025glassguidedlatentslot}
Krishnakant Singh, Simone Schaub-Meyer, and Stefan Roth.
\newblock Glass: Guided latent slot diffusion for object-centric learning,
  2025.

\bibitem[Wang et~al.(2004)Wang, Bovik, Sheikh, and Simoncelli]{wangssim}
Zhou Wang, A.C. Bovik, H.R. Sheikh, and E.P. Simoncelli.
\newblock Image quality assessment: from error visibility to structural
  similarity.
\newblock \emph{IEEE TIP}, 13\penalty0 (4):\penalty0 600--612, 2004.

\bibitem[Watters et~al.(2019)Watters, Matthey, Burgess, and
  Lerchner]{watters2019spatialbroadcastdecodersimple}
Nicholas Watters, Loic Matthey, Christopher~P. Burgess, and Alexander Lerchner.
\newblock Spatial broadcast decoder: A simple architecture for learning
  disentangled representations in vaes.
\newblock In \emph{ICLR}, 2019.

\bibitem[Wu et~al.(2023{\natexlab{a}})Wu, Dvornik, Greff,
  et~al.]{wu2023slotformerunsupervisedvisualdynamics}
Ziyi Wu, Nikita Dvornik, Klaus Greff, et~al.
\newblock Slotformer: Unsupervised visual dynamics simulation with
  object-centric models.
\newblock In \emph{ICLR}, 2023{\natexlab{a}}.

\bibitem[Wu et~al.(2023{\natexlab{b}})Wu, Hu, Lu,
  et~al.]{wu2023slotdiffusionobjectcentricgenerativemodeling}
Ziyi Wu, Jingyu Hu, Wuyue Lu, et~al.
\newblock Slotdiffusion: Object-centric generative modeling with diffusion
  models, 2023{\natexlab{b}}.

\bibitem[Yu et~al.(2025)Yu, Kwak, Jang, Jeong, Huang, Shin, and
  Xie]{yu2025representationalignmentgenerationtraining}
Sihyun Yu, Sangkyung Kwak, Huiwon Jang, Jongheon Jeong, Jonathan Huang, Jinwoo
  Shin, and Saining Xie.
\newblock Representation alignment for generation: Training diffusion
  transformers is easier than you think, 2025.

\bibitem[Zadaianchuk et~al.(2023)Zadaianchuk, Seitzer, and
  Martius]{zadaianchuk2023objectcentriclearningrealworldvideos}
Andrii Zadaianchuk, Maximilian Seitzer, and Georg Martius.
\newblock Object-centric learning for real-world videos by predicting temporal
  feature similarities, 2023.

\bibitem[Zhang et~al.(2018)Zhang, Isola, Efros, Shechtman, and
  Wang]{zhang2018unreasonableeffectivenessdeepfeatures}
Richard Zhang, Phillip Isola, Alexei~A. Efros, Eli Shechtman, and Oliver Wang.
\newblock The unreasonable effectiveness of deep features as a perceptual
  metric.
\newblock In \emph{CVPR}, 2018.

\bibitem[Zheng et~al.(2025)Zheng, Ma, Tong,
  et~al.]{zheng2025diffusiontransformersrepresentationautoencoders}
Boyang Zheng, Nanye Ma, Shengbang Tong, et~al.
\newblock Diffusion transformers with representation autoencoders, 2025.

\end{thebibliography}
}

\clearpage
\appendix
\section*{APPENDIX}
\renewcommand{\thesubsection}{A.\arabic{subsection}}
\setcounter{subsection}{0}

\subsection{Training and hyperparameter details}
Slot-RAE is designed to be trained with high computational efficiency, a significant advantage over heavier, pre-trained large-scale models based on Stable Diffusion. The complete training process is executed on a single NVIDIA A100 (80GB) GPU, avoiding the need for complex distributed multi-node setups.
\begin{itemize}
    \item \textbf{Hardware setup:} 1$\times$ NVIDIA A100 GPU (80GB VRAM), full-precision.
    \item \textbf{Duration:} 40 hours of continuous training (2 days). 500k steps.
    \item \textbf{Batch size:} An effective batch size of 64 is used.
    \item \textbf{Optimizer:} Adam optimizer is employed with learning rate scheduling.
    \item \textbf{Learning rate schedule:} We use a peak learning rate of $1\times 10^{-4}$ following an exponential decay schedule. To ensure stable early training dynamics, a linear warmup phase of 10,000 steps is applied.
\end{itemize}

\subsection{Architecture specifications}
The architecture of Slot-RAE is highly streamlined. Below are the exact specifications for the Base variant used in our primary evaluations:

\begin{itemize}
    \item \textbf{Feature Backbone (Frozen):} DINOv3-ViT-Base~\cite{Simeoni2025DINOv3}. The target feature dimension $d_{in}$ is $768$. We extract features from the final layer to capture high-level semantic information.
    \item \textbf{Slot Attention Module:} Configured with $K=7$ slots for the COCO dataset and $K=6$ for the PASCAL-VOC dataset, accounting for the average maximum number of distinct entities per scene. The slot dimension $d_{slot}$ is set to $768$, aligning with features dimensions. We use 3 iterative routing steps.
    \item \textbf{DiT (Diffusion Transformer Decoder):} 
        \begin{itemize}
            \item \textbf{Layers:} 12 standard Transformer blocks
            \item \textbf{Hidden Dimension:} $768$.
            \item \textbf{Attention Heads:} 12 heads for transformer path and 16 for additional wide DDT head~\cite{singh2026improvedbaselinesrepresentationautoencoders}.
        \end{itemize}
    \item \textbf{REPA Head:} A lightweight 2-layer Multi-Layer Perceptron (MLP). $\lambda$ loss parameter is set to 1.0.
\end{itemize}

\subsection{Additional Qualitative Results}

\begin{figure}[htbp]
    \centering
    \begin{tabularx}{\linewidth}{ *{4}{>{\centering\arraybackslash}X} }
        \textbf{Original} & \textbf{Stable-LSD} & \textbf{GLASS} & \textbf{Ours} \\
    \end{tabularx}
    \vspace{1mm}
    \includegraphics[width=\linewidth]{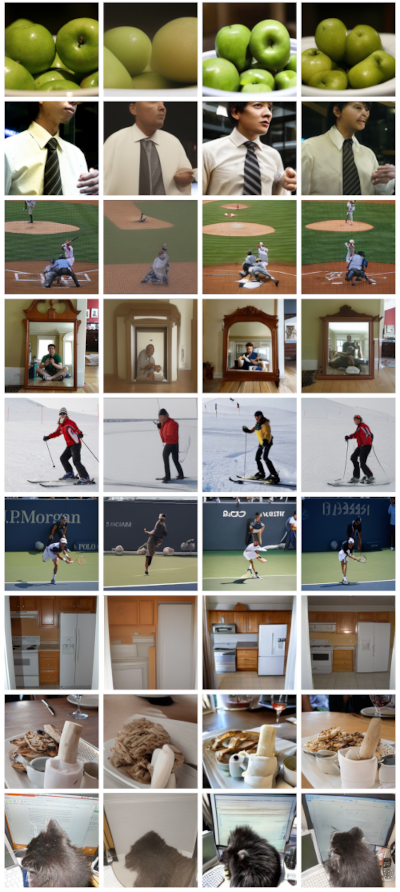}
    \caption{\textbf{Qualitative comparisons of image reconstruction} capabilities on the COCO dataset. From left to right: Original input, Stable-LSD baseline~\cite{jiang2023objectcentricslotdiffusion}, GLASS baseline~\cite{singh2025glassguidedlatentslot}, and our proposed Slot-RAE.}
    \label{fig:model_comparison_annex_1}
\end{figure}

In Figure~\ref{fig:model_comparison_annex_1}, we present an expanded qualitative analysis of image reconstruction capabilities across the COCO dataset. These examples showcase a wide variety of challenging domains and object interactions.

Consistent with the quantitative findings in the main text, these diverse and visually cluttered scenes visually confirm that our proposed Slot-RAE method better preserves fine-grained textural details, object boundaries, and global spatial coherence compared to existing baselines.

\begin{figure*}[htbp]
    \centering
    \begin{subfigure}[b]{0.49\textwidth}
        \includegraphics[width=\linewidth]{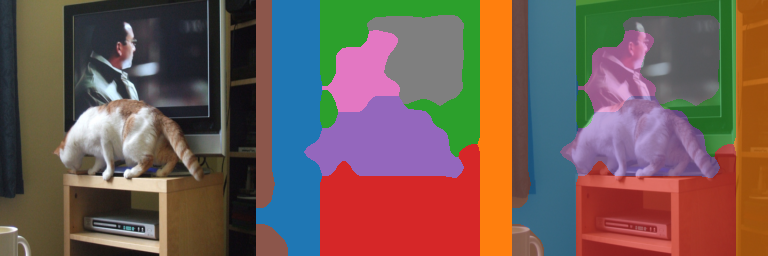}
    \end{subfigure}\hfill
    \begin{subfigure}[b]{0.49\textwidth}
        \includegraphics[width=\linewidth]{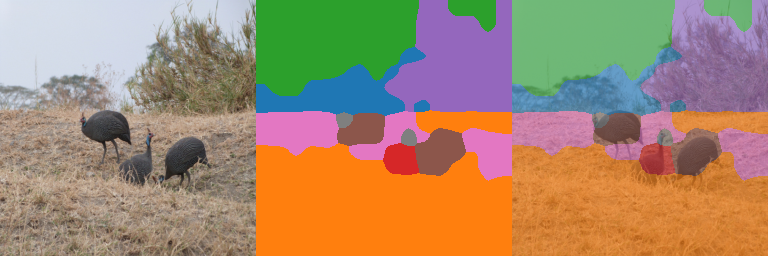}
    \end{subfigure}
    
    \vspace{2mm}
    
    \begin{subfigure}[b]{0.49\textwidth}
        \includegraphics[width=\linewidth]{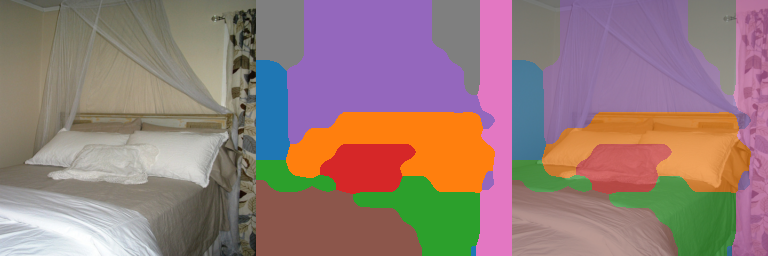}
    \end{subfigure}\hfill
    \begin{subfigure}[b]{0.49\textwidth}
        \includegraphics[width=\linewidth]{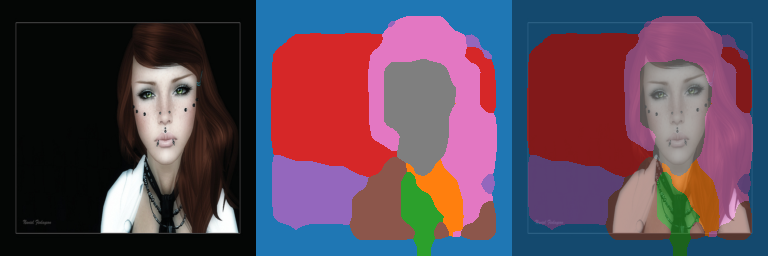}
    \end{subfigure}
    
    \vspace{2mm}
    
    \begin{subfigure}[b]{0.49\textwidth}
        \includegraphics[width=\linewidth]{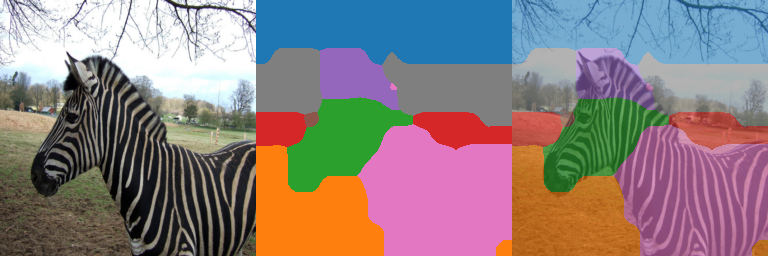}
    \end{subfigure}\hfill
    \begin{subfigure}[b]{0.49\textwidth}
        \includegraphics[width=\linewidth]{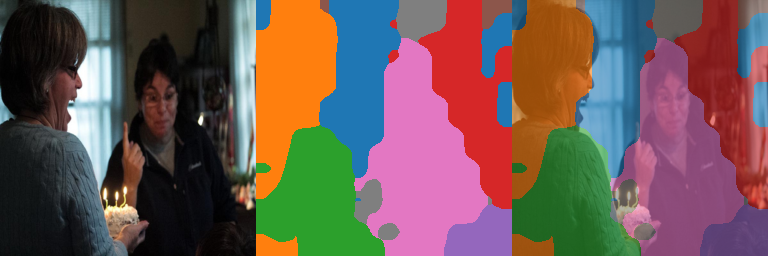}
    \end{subfigure}
    
    \vspace{2mm}
    
    \begin{subfigure}[b]{0.49\textwidth}
        \includegraphics[width=\linewidth]{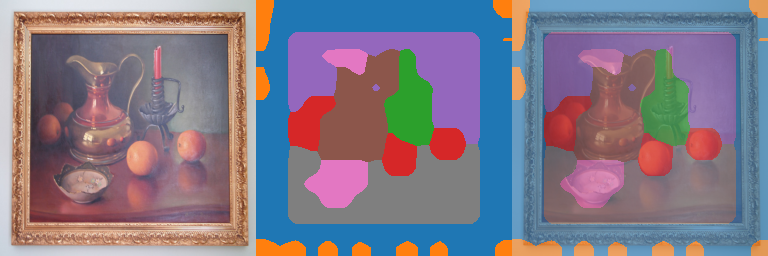}
    \end{subfigure}\hfill
    \begin{subfigure}[b]{0.49\textwidth}
        \includegraphics[width=\linewidth]{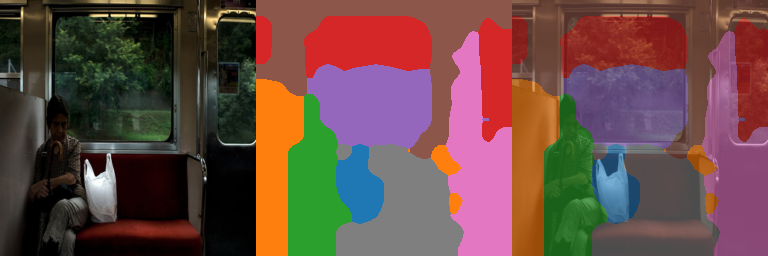}
    \end{subfigure}
    
    \vspace{2mm}

    \begin{subfigure}[b]{0.49\textwidth}
        \includegraphics[width=\linewidth]{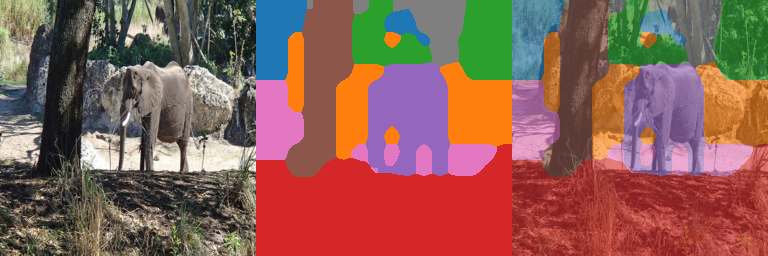}
    \end{subfigure}\hfill
    \begin{subfigure}[b]{0.49\textwidth}
        \includegraphics[width=\linewidth]{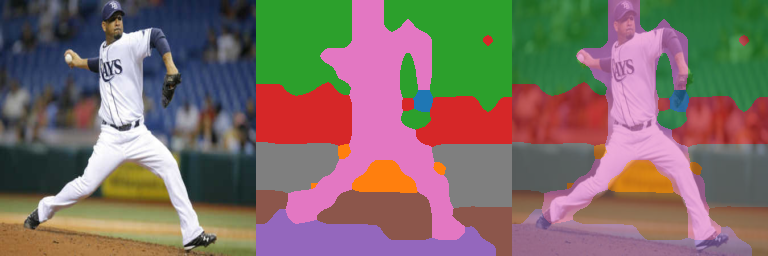} 
    \end{subfigure}

    \caption{\textbf{Additional qualitative results for unsupervised object discovery on the COCO validation dataset.} Each panel displays the original input image (left), the spatial attention masks derived from the learned object slots (center), and the overlaid segmentation (right). Slot-RAE effectively isolates distinct entities and backgrounds directly within the foundation model feature space without relying on external generative priors.}
    \label{fig:object_discovery_grid}
\end{figure*}

In addition to image reconstruction, Figure~\ref{fig:object_discovery_grid} provides further qualitative examples of unsupervised object discovery on the COCO validation set. These visualizations illustrate the spatial attention masks natively induced by our slot-attention bottleneck prior to the diffusion decoding process. As demonstrated, Slot-RAE successfully decomposes complex, cluttered scenes into discrete, semantically meaningful entities. Despite lacking explicit supervision or guidance from external pre-trained text models, our method effectively isolates distinct objects, delineates sharp boundaries, and separates foreground subjects from complex backgrounds.

\subsection{Ablation Studies}
\label{sec:ablat}
We conduct a series of comprehensive ablation experiments to identify the key factors contributing to Slot-RAE's performance. Unless otherwise specified, all ablations are performed on the COCO dataset using the same training budget and evaluation protocol as the main experiments. 

To provide a clear, holistic view of our architectural decisions, we aggregate the results of all ablation studies into a single summary table (\cref{tab:master_ablation}).

\begin{table*}[htbp]
\centering
\caption{\textbf{Comprehensive ablation studies on the COCO dataset.} We systematically isolate the impact of Decoder Scale, Foundation Model features, the REPA objective, Slot Dimensions, and Multi-Layer Integration. Instead of repeating our baseline configuration across sections, the best results (Slot-RAE Base with 768-dim slots) are detailed in the top shaded row. \textbf{Bold} indicates the best performance within the evaluated metrics, while \underline{underline} indicates the second best.}
\label{tab:master_ablation}
\resizebox{\textwidth}{!}{
\begin{tabular}{l|l ccccccc}
\toprule
\textbf{Ablation Category} & \textbf{Configuration / Variant} & \textbf{mBO$_i$} $\uparrow$ & \textbf{mBO$_c$} $\uparrow$ & \textbf{PSNR} $\uparrow$ & \textbf{SSIM} $\uparrow$ & \textbf{LPIPS} $\downarrow$ & \textbf{FID} $\downarrow$ \\
\midrule
\rowcolor{gray!20} \textbf{Best Results} & \textbf{Slot-RAE-B, DINOv3-B, REPA, 768-dim} & 32.3 & 38.1 & \textbf{13.93} & \textbf{0.33} & \textbf{0.46} & \textbf{10.11} \\
\midrule
\multirow{3}{*}{\textbf{(a) Decoder Scale}} 
& Slot-RAE-S (60M)  & 32.3 & 38.7 & 13.27 & 0.31 & 0.50 & 11.92 \\
& Slot-RAE-L (470M) & 32.5 & 38.3 & 13.50 & \underline{0.32} & 0.48 & 10.44 \\
& Slot-RAE-XL (839M)& 32.4 & 38.1 & 13.31 & \underline{0.32} & 0.49 & 10.61 \\
\midrule
\multirow{2}{*}{\textbf{(b) VFM Encoder}} 
& DINOv2-S & 29.4 & 34.3 & 12.83 & 0.29 & 0.51 & 10.45 \\
& DINOv2-B & 29.1 & 34.4 & 12.94 & 0.30 & 0.52 & 11.85 \\
& DINOv3-S & 31.4 & 37.3 & 13.40 & 0.31 & 0.49 & 10.31 \\
\midrule
\textbf{(c) REPA Head} 
& Without REPA (\xmark) & 31.7 & 37.9 & 13.53 & \underline{0.32} & 0.49 & 10.54 \\
\midrule
\multirow{3}{*}{\textbf{(d) Slot Dimension}} 
& 256 & \underline{32.8} & \underline{38.4} & 13.64 & \underline{0.32} & 0.48 & 10.52 \\
& 384 & 32.7 & 38.3 & \underline{13.77} & \underline{0.32} & \underline{0.47} & \underline{10.25} \\
& 512 & 32.4 & 38.1 & 13.57 & \underline{0.32} & 0.49 & 10.71 \\
\midrule
\multirow{4}{*}{\textbf{(e) Component Analysis}} 
& \makecell[l]{Slot Attn + DINOv2-B \\ (DINOSAUR + Transf.) No REPA (\xmark)} & \textbf{33.3} & \textbf{41.2} & $\emptyset$ & $\emptyset$ & $\emptyset$ & $\emptyset$ \\
& \quad + DiT Decoder & 29.1 & 34.4 & 12.94 & 0.30 & 0.52 & 11.85\\
& \quad + DINOv3-B & 31.7 & 37.9 & 13.53 & \underline{0.32} & 0.49 & 10.54 \\
\midrule
\textbf{(f) Multi-Layer Integration}
& Naïve MSL (Addition) & 29.3 & 33.8 & 10.96 & 0.27 & 0.64 & 46.39 \\
\bottomrule
\end{tabular}
}
\end{table*}

\subsubsection{Effect of Decoder Scale}
We first investigate how the capacity of the diffusion decoder impacts object discovery and reconstruction quality (Table~\ref{tab:master_ablation}a). We evaluate four Diffusion Transformer variants following the standard DiT scaling strategy: Small (S), Base (B), Large (L), and Extra-Large (XL), all trained using DINOv3-B features. Interestingly, while increasing decoder capacity from Small to Base improves instance-level object discovery ($\text{mBO}_i$) and overall feature synthesis, further scaling up to Large and XL results in diminishing returns. The XL variant, in particular, exhibits slight drops in both decomposition ($\text{mBO}_i$ of 32.4) and reconstruction quality. This behavior stands in contrast to traditional deterministic generative architectures, such as the original RAE framework~\cite{singh2026improvedbaselinesrepresentationautoencoders}, where scaling up decoder capacity consistently yields monotonic improvements in reconstruction fidelity. We hypothesize that in a slot-conditioned diffusion paradigm, excessively large decoders introduce a capacity mismatch. Specifically, scene decomposition primarily depends on the representational quality of the slot bottleneck; when the decoder becomes overly expressive, it begins to bypass the bottleneck's routing constraints. Instead of leveraging localized slot information, the over-parameterized decoder overpowers the slots by memorizing global layout contexts, ultimately degrading both strict object-centric routing and downstream reconstruction.

\subsubsection{Vision Foundation Model Encoder}
Next, we evaluate the influence of the underlying semantic feature space (Table~\ref{tab:master_ablation}b). We swap the foundation model features between DINOv2-S, DINOv2-B, DINOv3-S, and DINOv3-B while keeping the Slot-RAE-B architecture fixed. We observe that stronger foundation-model representations (DINOv3-B) significantly boost both instance and class-level decomposition metrics ($\text{mBO}_i$ and $\text{mBO}_c$) compared to older versions. This result ultimately supports our central hypothesis: generative object-centric learning scales directly with the quality of the underlying semantically structured feature spaces. It also follows findings on generation quality observed in~\cite{singh2026improvedbaselinesrepresentationautoencoders}.

\subsubsection{Effect of Representation Alignment (REPA)}
The Representation Alignment (REPA) objective was originally introduced to stabilize continuous feature-space diffusion training. In Table~\ref{tab:master_ablation}c, we assess its contribution in the unsupervised object-centric setting. Training Slot-RAE without REPA causes a degradation across all metrics. The inclusion of REPA consistently improves convergence stability and reconstruction quality. Notably, removing it causes a $0.6$ point drop in object discovery ($\text{mBO}_i$), suggesting that forcing the diffusion features to align directly with the VFM topological space encourages the slot bottleneck to capture more coherent, discrete object-level structures.

\subsubsection{Effect of Slot Dimensions}
As shown in Table~\ref{tab:master_ablation}d, we ablate the latent capacity of the slots by varying the dimension from 256 up to 768. We find that a smaller, more compressed slot dimension of 256 acts as a stronger informational bottleneck. This compression forces the model to factorize distinct objects more cleanly rather than redundantly distributing feature representations across multiple high-dimensional slots, leading to marginal but consistent improvements in discovery ($\text{mBO}_i$). However, scaling the slot dimensions up to 768 yields the absolute best overall reconstruction metrics across the board (e.g., reaching a PSNR of 13.93 and an FID of 10.11). This clearly highlights a trade-off where smaller dimensions prioritize strict object factorization, while heavily expanded dimensions excel at high-fidelity synthesis at the slight expense of discovery performance.

\subsubsection{Component Analysis}
Table~\ref{tab:master_ablation}e traces the progressive evolution of our architecture from a standard DINOSAUR baseline to the fully equipped Slot-RAE. While integrating the DiT decoder allows for generative synthesis, it initially causes a severe drop in decomposition metrics (e.g., $\text{mBO}_i$ dropping from 33.3 to 29.1) if improperly regularized. Upgrading to DINOv3 recovers some semantic capability, but the addition of the REPA objective is the critical piece that harmonizes the generative DiT with the VFM features. The results indicate that the largest generative gains arise from combining feature-space diffusion with representation alignment.

\subsubsection{Multi-layer Analysis \& Limitations}
While our primary architecture extracts object-centric representations from a single VFM layer (specifically the final layer, which contains the highest-level semantic information), we also conducted a probing experiment to assess the viability of multi-layer guidance. This exploration is motivated by RAEv2~\cite{singh2026improvedbaselinesrepresentationautoencoders}, which demonstrates superior representation quality through multi-layer feature aggregation, as well as the recent object-centric model MUFASA~\cite{bock2026mufasamultilayerframeworkslot}, which processes multi-level features using distinct, parallel slot-attention modules. 

To test whether Slot-RAE could benefit from such an integration, we evaluated a naïve multi-layer variant (Table~\ref{tab:master_ablation}f) that adopts the RAEv2 addition setup (MSL) by simply summing features across layers without introducing heavy parallel attention modules. Unfortunately, this naive formulation drastically degrades both decomposition and reconstruction quality, evidenced by the severe drop in PSNR to 10.96 and an extreme FID spike to 46.39. This finding implies that while foundation model feature hierarchies contain rich, complementary semantics (low-level edges vs. high-level classes), simply aggregating them prior to a unified slot-attention bottleneck introduces severe optimization conflicts and blurs object boundaries. Developing a more sophisticated, yet computationally efficient mechanism to integrate multi-layer VFM guidance into the Slot-RAE framework remains an open direction for future work.

\subsection{Extended Visualizations of Composition and Subtraction}

To further validate the generalization and robustness of our semantic feature space, we provide extended qualitative results for zero-shot object manipulation. 

\begin{figure*}[htbp]
    \centering
    
    \begin{subfigure}[b]{0.7\linewidth}
        \centering
        \includegraphics[width=\linewidth]{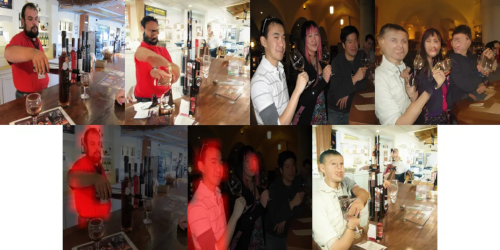}
        \caption{}
    \end{subfigure}
    
    \vspace{0.5em}
    
    \begin{subfigure}[b]{0.7\linewidth}
        \centering
        \includegraphics[width=\linewidth]{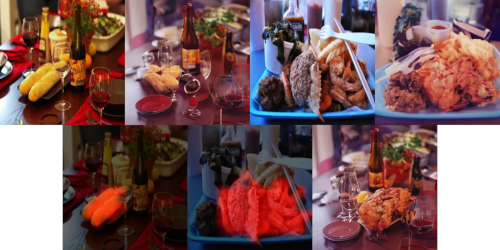}
        \caption{}
    \end{subfigure}
    
    \vspace{0.5em}
    
    \begin{subfigure}[b]{0.7\linewidth}
        \centering
        \includegraphics[width=\linewidth]{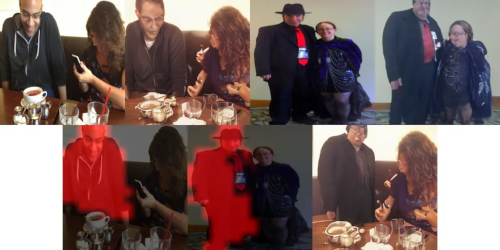}
        \caption{}
    \end{subfigure}

    \caption{\textbf{Additional results for zero-shot composition.} We demonstrate the model's ability to seamlessly blend slots from diverse and complex scenes. For each example \textbf{(a-c)}, the top row displays Source Image A, Reconstruction A, Source Image B, and Reconstruction B. The bottom row illustrates the composition process, where a target slot from Image A is swapped with a selected slot from Image B to generate the final composed scene (Image A + Slot from B). The lighting, shadows, and object interactions are naturally resolved in the composed outputs, preserving the structural integrity of both the target environment and the injected semantic objects.}
    \label{fig:extended_composition}
\end{figure*}

Figure~\ref{fig:extended_composition} illustrates additional examples of semantic composition. The model effectively resolves challenging physical interactions, adjusting lighting, shadows, and occlusions to ensure the injected elements are naturally grounded within the new target environment.

\begin{figure*}[htbp]
    \centering
    
    \begin{subfigure}[b]{0.8\linewidth}
        \centering
        \includegraphics[width=\linewidth]{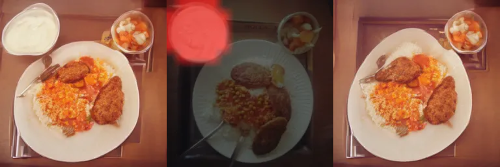}
    \end{subfigure}
    
    \vspace{0.5em}
    
    \begin{subfigure}[b]{0.8\linewidth}
        \centering
        \includegraphics[width=\linewidth]{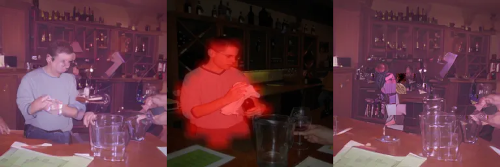}
    \end{subfigure}
    
    \vspace{0.5em}
    
    \begin{subfigure}[b]{0.8\linewidth}
        \centering
        \includegraphics[width=\linewidth]{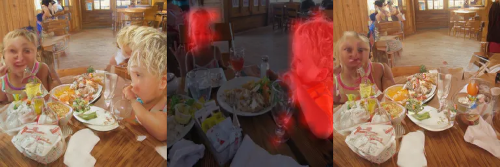}
    \end{subfigure}
    
    \vspace{0.5em}
    
    \begin{subfigure}[b]{0.8\linewidth}
        \centering
        \includegraphics[width=\linewidth]{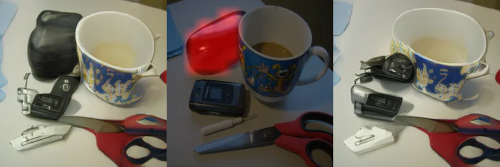}
    \end{subfigure}

    \caption{\textbf{Additional results for zero-shot subtraction.} The model effectively localizes target semantic concepts (highlighted in red) and removes them from the scene. The occluded backgrounds are realistically inpainted, maintaining global context and structural coherence even in highly cluttered environments such as dining tables and complex indoor settings.}
    \label{fig:extended_subtraction}
\end{figure*}

Figure~\ref{fig:extended_subtraction} showcases further results for zero-shot slot subtraction. The model is tasked with localizing specific semantic concepts (indicated by the red attention masks) and entirely removing them from the scene. Even in dense and highly cluttered environments Slot-RAE successfully preserves the global context. It naturally inpaints the heavily occluded regions, maintaining the structural coherence of the underlying geometry and background textures.

\subsection{Limitations and Failure Cases}

\begin{figure}[htbp]
    \centering
    
    \begin{subfigure}[b]{\linewidth}
        \centering
        \includegraphics[width=\linewidth]{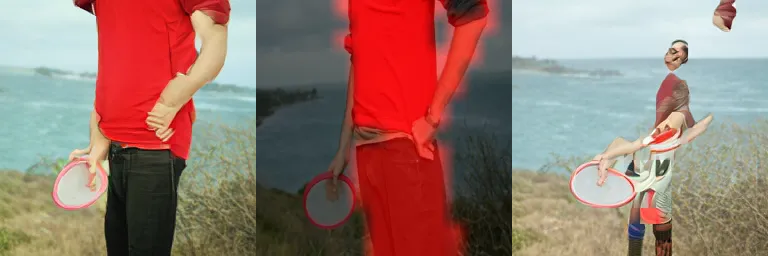}
    \end{subfigure}
    
    \vspace{0.5em}
    
    \begin{subfigure}[b]{\linewidth}
        \centering
        \includegraphics[width=\linewidth]{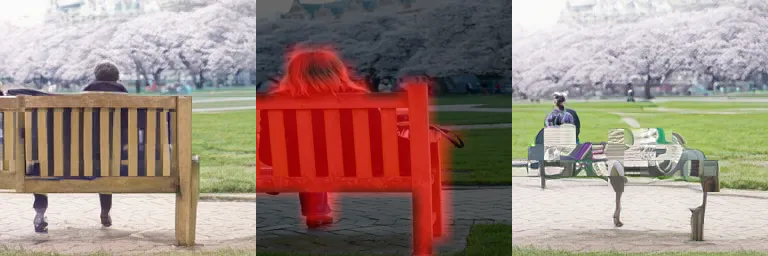}
    \end{subfigure}
    
    \vspace{0.5em}
    
    \begin{subfigure}[b]{\linewidth}
        \centering
        \includegraphics[width=\linewidth]{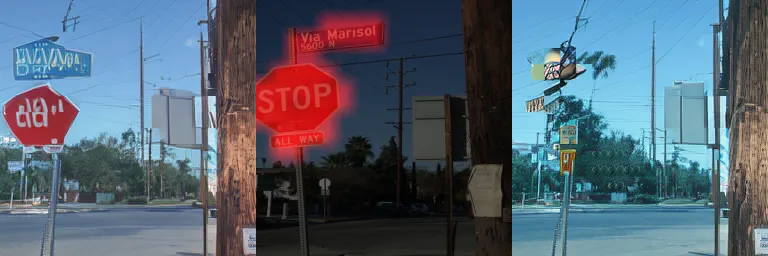}
    \end{subfigure}

    \caption{\textbf{Failure cases of zero-shot slot subtraction.} While our model effectively localizes and manipulates semantic concepts, removing slots in highly complex scenes can expose limitations in background generation. \textit{Top:} The model successfully removes the primary subject but fails to disentangle highly correlated elements, leaving behind floating hands and distorted textures. \textit{Middle:} Subtracting a heavily occluding subject disrupts the underlying geometry of the scene, causing the structural collapse of the wooden bench. \textit{Bottom:} In dense environments with overlapping objects, removing a target slot (street signs) can leave floating structural artifacts and incomplete sky inpainting.}
    \label{fig:failure_cases}
\end{figure}

While Slot-RAE demonstrates strong zero-shot compositional capabilities without relying on Classifier-Free Guidance, operating purely in a compressed semantic feature space occasionally presents challenges, particularly during slot subtraction in highly complex scenes. As illustrated in Figure~\ref{fig:failure_cases}, these failure cases primarily stem from ambiguous background priors. 

First, highly correlated objects can become entangled within the same slot representation. When attempting to subtract the primary subject, the model may fail to disentangle these elements entirely, inadvertently leaving behind floating appendages or distorted textures (Figure~\ref{fig:failure_cases}, Top). Second, removing a heavily occluding object requires the model to hallucinate the missing background geometry. In dense or highly structured environments, this localized inpainting can fail to preserve global structural integrity, leading to the collapse of underlying objects, such as the warped wooden bench (Figure~\ref{fig:failure_cases}, Middle), or leaving behind floating artifacts and incomplete sky generation (Figure~\ref{fig:failure_cases}, Bottom). Incorporating explicit depth priors or stronger geometric constraints to guide the DiT during zero-shot subtraction remains a promising direction for future work.

\end{document}